\documentclass[letterpaper, 10 pt, conference]{ieeeconf}  

\IEEEoverridecommandlockouts                              

\usepackage[T1]{fontenc}

\usepackage{amsfonts}       
\usepackage{nicefrac}       
\usepackage{subfig}			
\usepackage{booktabs}	
\usepackage[linesnumbered,ruled,vlined]{algorithm2e}
\usepackage{xcolor}
\usepackage[normalem]{ulem} 
\usepackage{mathtools}
\usepackage[percent]{overpic} 
\usepackage{flushend}
\usepackage{comment}
\usepackage[noadjust]{cite}

\definecolor{black}{rgb}{0, 0, 0}
\definecolor{red}{rgb}{0.9, 0, 0}
\definecolor{green}{rgb}{0, 0.6, 0}
\definecolor{blue}{rgb}{0, 0, 0.9}
\definecolor{grey}{rgb}{0.52, 0.52, 0.51}

\newcommand{\RED}[1]{\textcolor{red}{#1}}

\newcommand{\BLACK}[1]{\textcolor{black}{#1}}

\newcommand{\rev}[1]{\BLACK{#1}}

\newcommand{\mde}[0]{\texttt{MDE}}

\newcommand{\textt}[1]{\scalebox{.85}[1.0]{\texttt{#1}}}
\newcommand{\captionimg}[1]{\caption{\small #1}}

\SetKwInOut{KwParameter}{Parameters}
\SetKw{KwContinue}{continue}
\SetKw{KwBreak}{break}
\SetKw{KwInterval}{in}
\SetKw{KwOr}{or}
\SetKw{KwAnd}{and}
\SetKwFunction{unif}{unif}

\newcommand{\fillbox}[3]
{\bgroup
  \dimen1=#1\relax
  \dimen2=#2\relax
  \sbox0{\includegraphics[width=#1]{#3}}%
  \ifdim\ht0>\dimen2
    \dimen0=\dimexpr \ht0-\dimen2\relax
    \adjustbox{clip=true,trim=0pt 0.5\dimen0 0pt 0.5\dimen0}{\usebox0}%
  \else
    \sbox0{\includegraphics[height=#2]{#3}}%
    \ifdim\wd0>\dimen1
      \dimen0=\dimexpr \wd0-\dimen1\relax
      \adjustbox{clip=true,trim=0.5\dimen0 0pt 0.5\dimen0 0pt}{\usebox0}%
    \else
      \usebox0
    \fi
  \fi
\egroup}

\DeclareMathOperator*{\subjectto}{subject\;to}
\DeclareMathOperator*{\minimize}{minimize}

\DeclareMathOperator*{\argmax}{argmax}

\usepackage[export]{adjustbox}

\title{\LARGE \bf \rev{Uncertainty-aware Planning with Inaccurate Models for    \\ Robotized Liquid Handling}} 

\author{Marco Faroni, Carlo Odesco, Andrea M. Zanchettin, Paolo Rocco
\thanks{This study was partially carried out within the MICS (Made in Italy – Circular and Sustainable) Extended Partnership and received funding from Next-Generation EU (Italian PNRR – M4 C2, Invest 1.3 – D.D. 1551.11-10-2022, PE00000004). CUP MICS D43C22003120001.}
\thanks{
The authors are with the Department of Electronics, Information and Bioengineering, Politecnico di Milano, Italy. {\tt\footnotesize marco.faroni@polimi.it}
}
}

\begin{document}

\maketitle
\thispagestyle{empty}
\pagestyle{empty}

\begin{abstract}
Physics-based simulations and learning-based models are vital for complex robotics tasks like deformable object manipulation and liquid handling. 
However, these models often struggle with accuracy due to epistemic uncertainty or the sim-to-real gap. 
For instance, accurately pouring liquid from one container to another poses challenges, particularly when models are trained on limited demonstrations and may perform poorly in novel situations. 
This paper proposes an uncertainty-aware Monte Carlo Tree Search (MCTS) algorithm designed to mitigate these inaccuracies. 
By incorporating estimates of model uncertainty, the proposed MCTS strategy biases the search towards actions with lower predicted uncertainty. This approach enhances the reliability of planning under uncertain conditions. 
Applied to a liquid pouring task, our method demonstrates improved success rates even with models trained on minimal data, outperforming traditional methods and showcasing its potential for robust decision-making in robotics.
\end{abstract}

\section{Introduction}

\begin{figure}[tpb]
	\centering
	\includegraphics[trim = 2cm 3cm 28.0cm 0cm, clip, angle=0, width=0.9\columnwidth]{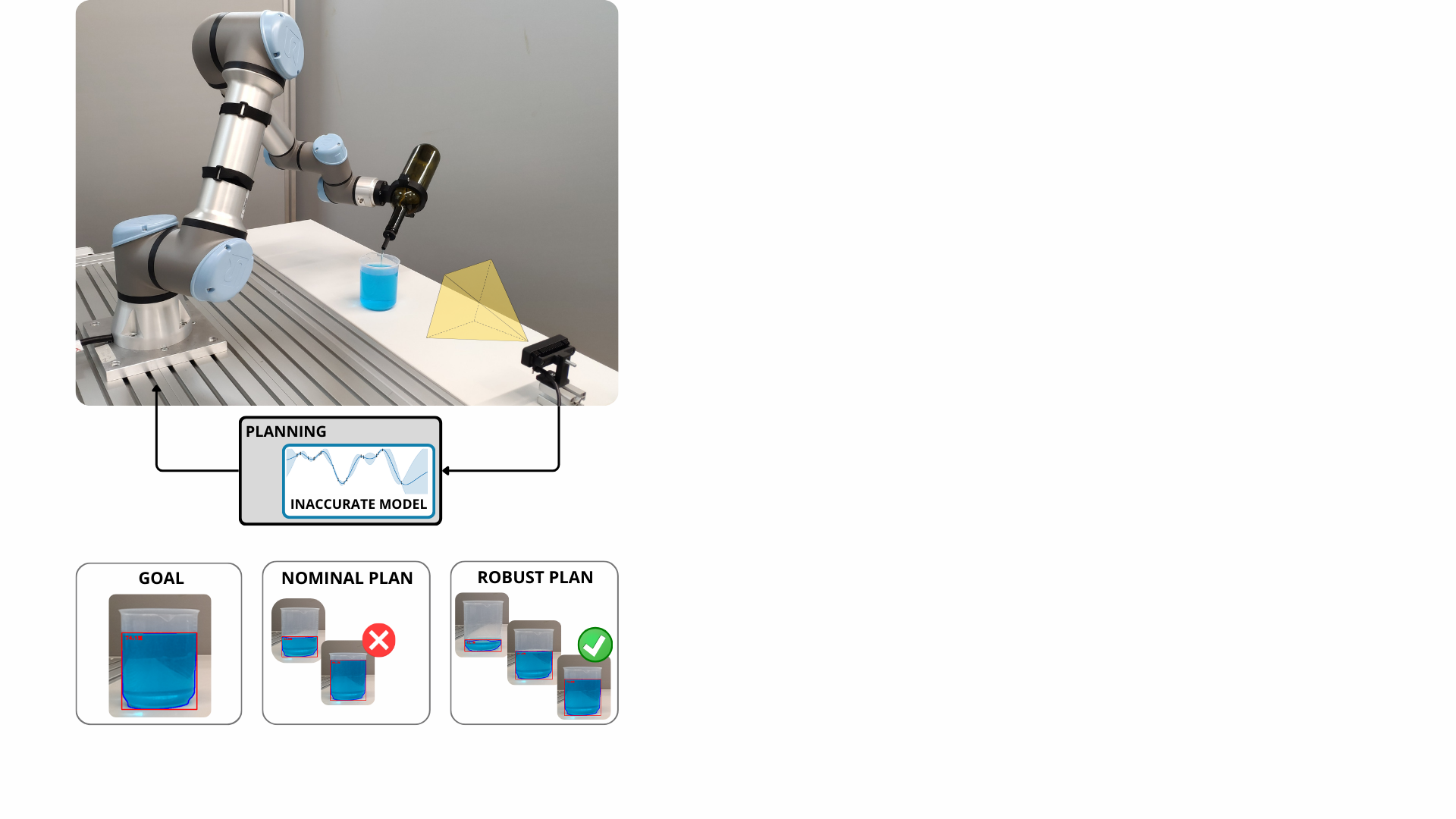}
	\captionimg{A robotized pouring task: the robot arm shall pour liquid from one container to another until it reaches a reference level.}
	\label{fig: setup}
 \vspace{-0.7cm}
\end{figure}

Physics-based simulations and learning-based models are extensively used in robotics to perform complex tasks such as deformable object manipulation  \cite{Lippi:visual-action-planning,NICOLA2024102630,Mitrano:science-robotics,corl-planning-with-spatial-temporal-abstraction,Mitrano:focus-adaptation}, contact-rich manipulation \cite{Kromere:contact-rich-manipulation,brock-contact-based-rrt,motion-planning-sliding}, control of soft robots \cite{garabini:motion-planning-soft-robot,marcucci2020parametric}, and liquid handling \cite{fluid-manipulation,coffee-steering}.
These models are often inaccurate in predicting the outcome of actions (e.g., because of the epistemic uncertainty of learned models or the sim-to-real gap of physics simulators). 
A notable example is the liquid handling scenario in Fig. \ref{fig: setup}: A robot has to pour liquid from one container to another until it reaches a certain level. 
Ideally, one could accurately model the quantity of poured liquid based on the robot's tool motion (e.g., via fluid simulation), but this requires a deep understanding of the geometry and dynamics properties of the problem or the collection of large experimental datasets. 
Suppose a pouring model is built from a handful of pouring demonstrations. 
The model will be very inaccurate, especially for states and actions that are far from those seen in the training set. 

If an estimate of the inaccuracy of the model predictions is available, this information can be embedded in the planning algorithm to avoid inaccurate states and actions \cite{Mitrano:science-robotics, Lagrassa:preconditions-mde-liquid}. 
Interestingly, machine learning techniques commonly used for dynamics modeling in robotics come with an estimate of the model epistemic uncertainty at a given point. 
For example, the variance of a Gaussian Process (GPs) is often seen as a proxy of the prediction uncertainty. 
Similarly, the variance of the predictions of Neural Network ensembles measures the consensus of the networks: if the variance is large, the average prediction is likely to be inaccurate.

This work proposes an uncertainty-aware Monte Carlo Tree Search (MCTS) algorithm that uses such uncertainty information to cope with model inaccuracy. 
\rev{We use the uncertainty value of each prediction to inform the MCTS search and bias it towards transitions with small uncertainty.}
As a result, the planner is likely to choose accurate states and actions, leading to more accurate task executions.

We demonstrate the approach with a liquid pouring problem. 
To do so, we model pourings with Gaussian Process Regression (GPR) and use the variance of GPs as an accuracy index for the model predictions. 
Then, we use the uncertainty-aware MCTS to choose the sequence of pouring actions to reach a given filling level.
We show that our approach performs pouring tasks even when the model uncertainty is very large (e.g., a GP with a handful of points), leading to a much higher success rate than the baselines.

\section{Related Works}\label{sec:related-works}

Planning under uncertainty gathers all those planning methods that explicitly reason about the difference between the model and the true system. 
The goal of these methods is to find a sequence of actions that brings the system from an initial to a final state even though the model and the true dynamics differ. 

If such a difference is due to the inherent randomness of the system (i.e., aleatoric uncertainty), a common approach is to propagate the uncertainty on the search tree. 
The planner will search for a robust solution that reaches the goal despite the transitions' randomness \cite{planning-belief-space, planning-belief-2022, Pavone:robust-rrt}. 
If the model error is caused by the limited dataset or the lack of expressivity of the model, we are dealing with epistemic uncertainty.
This last case is less studied and is the focus of this work. 

In the realm of sampling-based motion planning, existing approaches estimate a Model Deviation Estimate (\mde) predicting the accuracy of a transition and use it to bias the search. 
Mitrano et al. \cite{Mitrano:science-robotics} train a classifier to deem state-and-action pairs accurate or inaccurate during the search. 
Inaccurate transitions are not added to the tree in a sampling-based path planner.
Faroni et al. \cite{Faroni_RAL2023} use the \mde\ as a cost function within an asymptotically optimal planner so that the resulting path should minimize the cumulative model mismatch. 
Other approaches limit the search space to a trusted region after an offline analysis of the model performance \cite{Glen:trusted-domain}.
All these approaches rely on a computationally expensive training phase to learn which actions can be trusted and use such information in a general motion planning algorithm.
Considering graph-based search, Vemula et al. \cite{CMAX, CMAX++} use A$^*$ to plan robot movements and, after the execution of each action, assign a high cost to transitions whose deviation from the model is larger than a given threshold.

MCTS is becoming popular in robotics planning.
Recent applications include path planning with Partially Observable Markov Decision Processes (POMDP) \cite{mcts_jan_peters,mcts_path_planning_frontiers}, multi-robot task allocation \cite{mcts_task_allocation} and path planning \cite{mcts_hrc}, and rearrangement planning \cite{mcts_rearrangement}.
Existing works typically leverage MCTS to account for multi-agent decision-making, by modeling the multi-agency as a POMDP and using MCTS to solve the resulting problem.
Unlike previous works, ours uses MCTS to explicitly reason through model uncertainty in the context of robotics planning. 
Compared to existing sampling-based and graph-based planners, MCTS can be easily extended to account for uncertainty \cite{ua_mcts}.


Specific to robotized liquid pouring, most existing works focus on accurate modeling of the liquids' dynamics via deep learning or fluid simulation.
For example, Pan et al. \cite{manocha_simulation_pouring} combined a fine-grained fluid simulation with an optimization-based planner to achieve the pouring task. 
The motion planning problem was formulated as a continuous numerical optimization problem in the high-dimensional robot trajectory space, following an objective function that accounts for object avoidance, smoothness, and a target goal that ensures fluid particles enter the target container.
Similarly, Schober et al. \cite{schober2025vision} segment the image of the container, simulate possible pourings and select the best action to perform in the real-world. 
Zhang et al. \cite{zhang2024oneshot} used imitation learning to learn pouring tasks from human demonstrations. 
They devised a one-shot domain adaptation technique to reduce the sim-to-real gap between simulations and real-world environments. 
This led to a higher success rate in execution compared to non-adaptive methods.
Babaians et al. \cite{babaians2022pournet_rl} explored a deep reinforcement learning approach: the policy is learned in a simulated environment modeling the interaction between liquids and robots.
The main drawback of all these methods is that they rely on the availability of accurate models, whose computation requires a large data collection or long computational times.
We do not rely on precise modeling of the liquid dynamics. 
On the contrary, we aim to leverage the uncertainty of the model owing to limited data availability and show that the robot is still able to perform tasks accurately.

Other than planning-based methods, the most common approach to robotized pouring uses feedback control to continuously monitor the liquid level in the output container and regulate the pouring rate accordingly \cite{schneck2017visual,dong2019pid_pouring,do2019accurate_pouring}. 
This approach can achieve very accurate results but assumes the liquid's level is measured at a high sampling rate and without interruptions. 
On the contrary, we measure the liquid's level only between discrete pouring actions, thus performing each action in open loop. 
Therefore, our assumption is less restrictive, especially when camera occlusions may happen or the robot has an in-hand camera (not allowing for simultaneous action and perception).

\section{Problem Statement}\label{sec:problem-statement}

Consider a dynamical system, $x_{k+1} = f(x_k,u_k)$ where $f: X\times U \rightarrow X$ and $X$ and $U$ are the state space and the action space, respectively. 
Ideally, we aim to find a sequence of actions, $u=(u_0,\dots,u_N)$, $N\in \mathbb N$, that brings the system from its initial state, $x_{\mathrm{start}}$, to a desired goal set, $X_{\mathrm{goal}}$. 
Possibly, we aim to minimize the number of actions to reach the goal. 
We can write this problem as follows:
\begin{equation}
\label{eq: problem-ideal}
\begin{aligned}
    &\minimize_{u \in U} & & N \\
    &\subjectto  & & x_N \in X_{\mathrm{goal}}\\
    & & & x_0 = x_{\mathrm{start}}\\
    & & & x_{k+1} = f(x_k,u_k)
\end{aligned}
\end{equation}

In real-world problems, $f$ is approximated by a dynamical model. 
We consider the deterministic model $\hat{f}: X\times U \rightarrow X$ that approximates $f$, and define a Model Deviation Estimate, $\mde(x_k,u_k) \propto || f(x_k,u_k) - \hat{f}(x_k, u_k)  ||^2$, which outputs an estimate of the model mismatch when action $u_k$ is applied from state $x_k$.
The model deviation estimate can be derived from real data \cite{Mitrano:science-robotics}, analytically \cite{Faroni_RAL2023} or correlated to the model epistemic uncertainty.

Under the assumption that $x$ can be measured without uncertainty (i.e., $x_{\mathrm{start}}$ is known), we can solve the following approximation of \eqref{eq: problem-ideal}:
\begin{equation}
\label{eq: problem-real}
\begin{aligned}
    &\minimize_{u\in U} & & N \\
    &\subjectto  & & \hat{x}_N \in X_{\mathrm{goal}}\\
    & & & \hat{x}_0 = x_{\mathrm{start}}\\
    & & & \hat{x}_{k+1} = \hat{f}(\hat{x}_k,u_k)
\end{aligned}
\end{equation}
Problem \eqref{eq: problem-real} is similar to the standard formulation of a motion planning problem \cite{lavalle2006planning}. 
Motion planners usually find the whole solution to \eqref{eq: problem-real} beforehand and apply the entire sequence, $u$, in an open-loop fashion. 
Because we follow an MCTS approach, we apply one action at a time and repeat the search after updating $x_{\mathrm{start}}$ with the observed state.

Problem \eqref{eq: problem-real} does not use $\mde$, meaning that its solution might contain state-and-action pairs on which $\hat{f}$ makes inaccurate predictions.

Our goal is to overcome this issue by finding a solution whose state-and-action pairs have small \mde\ values. 
We do so by combining the GPs' variance with an MCTS algorithm to bias the search toward states and actions corresponding to low epistemic uncertainty.

\section{Method} \label{sec:method}
We use GPs' variance to inform the search of the proposed uncertainty-aware MCTS algorithm. 
This section describes how we address \eqref{eq: problem-real} within the MCTS framework, combine it with the \mde\, and provide insights on how GPR is used to compute the \mde.

\subsection{Uncertainty-aware Monte Carlo Tree Search} \label{sec: ua-mcts}

MCTS is a search algorithm that uses experience-driven heuristics to bias the tree growth. 
The algorithm relies on the definition of three functions:
\begin{itemize}
    \item a function returning all legal actions that can be taken at state $x_k$.
    \item a terminal-state checker returning whether a state is terminal (e.g., a goal or a state from which it is impossible to recover).
    \item a reward function, $r(x_k,u_k)$, which associates a state-and-action pair with a scalar reward.
\end{itemize}
The definition of these functions is problem-dependent. We will give examples for liquid pouring in Sec. \ref{sec: case-study}.

MCTS consists of four main phases, as shown in Alg.~\ref{alg: mcts}. 
Starting from the root node, it descends the tree by applying a \textbf{selection} rule until it reaches a leaf node. 
The leaf node undergoes an \textbf{expansion} procedure: all the actions available from that node are applied and the resulting nodes are appended to the tree. 
The algorithm picks one of the new nodes and applies random legal actions until it reaches a terminal state (\textbf{simulation}). 
At this point, the algorithm computes the reward and uses \textbf{backpropagation} to update the reward belief of all nodes on the tree walk from the root to the terminal state. 
The procedure repeats for a number of iterations and, finally, returns the most visited node.

\begin{algorithm}[tpb]
\DontPrintSemicolon
\SetKwFunction{MCTS}{MCTS}
\SetKwFunction{SELECT}{SELECT}
\SetKwFunction{EXPAND}{EXPAND}
\SetKwFunction{SIMULATE}{SIMULATE}
\SetKwFunction{BACKPROPAGATE}{BACKPROPAGATE}
\SetKwProg{Fn}{Function}{:}{}  
  \Fn{\MCTS{$x_0$}}{
    create a root node $v_0$ with state $v_0.x \leftarrow x_0$\;
    \For{$N_I$ iterations}{
        $v_s \leftarrow$ \SELECT{$v_0$}\;
        \If{$N(v_s)>0$}{
            $v_s \leftarrow$ \EXPAND{$v_s$}\;
        }
        $\mathrm{reward} \leftarrow$ \SIMULATE{$v_s.x$}\;
        \BACKPROPAGATE{$v_s,\mathrm{reward}$}\;
    }
    $v_{\mathrm{best}} \leftarrow$ choose the most visited child of $v_0$\;
    \KwRet action that generated $v_{\mathrm{best}}$\;
  }
\caption{MCTS algorithm. Adapted from \cite{ua_mcts}.}
\label{alg: mcts}
\end{algorithm}

We embed the \mde\ function in all these steps using a variant of the MCTS proposed in \cite{ua_mcts}.
\cite{ua_mcts} modifies the selection, expansion, backpropagation, and simulation phases to account for action uncertainty. 
The ablation studies in \cite{ua_mcts} demonstrated that the uncertainty-aware variations of the selection and expansion phases are those that impact the search the most. 
In contrast, backpropagation and simulation do not significantly affect the results. 
For this reason, we propose the following uncertainty-aware selection and expansion functions:
\subsubsection{Selection}
    Selection is applied to already visited nodes. The goal is to choose promising nodes from which to expand the tree. MCTS uses UCT, an extension of Upper Confidence Bound (UCB) to trees.
    Given a node in the tree, $v$, and a set of children, $\mathcal{V}$, UCT chooses the next node, $v_{\mathrm{next}}$, as:
    \begin{equation}
    \label{eq: uct}
        v_{\mathrm{next}} = \argmax_{v_i \in \mathcal{V}} \frac{Q(v_i)}{N(v_i)} + c \sqrt{\frac{\ln{N(v)}}{N(v_i)}} 
    \end{equation}    
    where  $Q(v)$ returns the sum of rewards observed by $v$ and $N(v)$ returns how many times $v$ has been visited so far; $c>0$ is a user-defined constant.
    The first term in \eqref{eq: uct} is an estimate of the average reward of $v$; the second one is inversely proportional to the number of times $v$ has been selected.
    As a consequence, UCT trades off exploration of less visited nodes and exploitation of nodes that returned a high reward in the past.
    We modify the UCT rule to make it uncertainty-aware as follows:
    \begin{equation}
    \label{eq: uct-ua}
        v_{\mathrm{next}} = \argmax_{v_i \in \mathcal{V}}  \Bigg( \frac{Q(v_i)}{N(v_i)} + c \sqrt{\frac{\ln{N(v)}}{N(v_i)}} \Bigg) \cdot (1-\delta_i)
    \end{equation}    
    where  
    \[
        \delta_i = \frac{e^{\mde(v_i)/\tau}}{\sum_{v_j \in \mathcal{V}} e^{\mde(v_j)/\tau}}
    \]
    is the softmax function with temperature parameter $\tau$ and $\mde(v)$ refers to the \mde\ function evaluated in the state and action that generated $v$.
    In this way, the selection phase is biased towards nodes with a low {\mde} value.
    The modified selection function is in Alg. \ref{alg: selection}.
    %

\subsubsection{Expansion}
    
    Expansion means generating the children of a leaf node, $v$, by applying a legal action $a_i$ from the node state, $v.x$. 
    The expansion function can be modified to favor certain children or to limit the branching factor (for computational efficiency). In our algorithm, we randomly discard children with a probability proportional to their \mde\ value.
    In detail, we denote by $\theta$ the average \mde\ of all children of the node and discard a child $v_i$ with a probability given by the following sigmoid function:
    \begin{equation}
        p(v_i) = \frac{1}{1+e^{h(\mde(v_i) - \theta) }}
    \end{equation}
    where $h$ is the steepness of the sigmoid.
    The result is that children with \mde\ values greater than the average will have a low probability of being expanded.
    The modified expansion function is in Alg. \ref{alg: expansion}.

\emph{Remark 1 (Effect of parameters $h$ and $\tau$): }
The selection and expansion rules depend on the sigmoid steepness, $h$, and the softmax temperature, $\tau$. 
$h$ and $\tau$ determine how biased the algorithm is toward low-uncertainty actions. 
In particular, large values of $h$ increase the likelihood of rejecting actions whose uncertainty is higher than the average uncertainty $\theta$. 
Similarly, small values of $\tau$ result in higher selection probabilities for low-uncertainty actions.

\rev{
\emph{Remark 2 (Continuous action-spaces): }
    MCTS is naturally suited for discrete action spaces. 
    For this reason, our approach was formulated for such a case. 
    The approach extends to sampling-based continuous-space MCTS variants (e.g., \emph{progressive widening} \cite{progressive-widening}), which sample new actions from the continuous space and apply selection and expansion to a growing set of actions. 
    In this case, we can simply replace the selection and expansion functions with the proposed uncertainty-aware versions.
}

\subsection{Modeling uncertainty} \label{sec: modeling-uncertainty}

Previous works computed an \mde\ function analytically \cite{Faroni_RAL2023}\cite{Faroni_ICRA24} or from a dataset of robot trajectory executions \cite{Mitrano:science-robotics}\cite{Lagrassa:preconditions-mde-liquid}. 
Our uncertainty-aware MCTS is independent of the approach used to retrieve the \mde. 
Nonetheless, it is worth noting that some of the most widespread modeling techniques in robotics (e.g., GPs and neural-network ensembles) come with an estimate of the epistemic uncertainty of the model's predictions. 
\rev{The validation of our approach uses GPR to model the system dynamics. 
For this reason, we give insights on the usage of GPR to this purpose.} 

GPR is a non-parametric regression approach. 
It uses GPs to infer a Gaussian distribution for a new point given a set of observations and a kernel function. 

Let $y=\{y_1, \dots, y_H\}$ be the set of observed values corresponding to features vectors $z=\{z_1, \dots, z_H\}$ and $z'=\{z'_1, \dots, z'_H\}$ the set of new feature vectors for inference.
In a GP, the joint distribution of observed and predicted values is a Gaussian distribution:
\begin{equation}
    \begin{pmatrix}
    y \\
    y'
    \end{pmatrix}\sim \mathcal{N}\left(\begin{pmatrix}
    0 \\
    0
    \end{pmatrix},\begin{pmatrix}
    k(z,z) & k(z,z') \\
    k(z',z) & k(z',z')
    \end{pmatrix}\right)
\end{equation}
from which it is possible to infer $y'$ as a Gaussian distribution with mean $\mu$ and variance $\sigma^2$. 

We can use GPR to model our system's dynamics, $f$. 
By denoting the feature vector by $z_k = (x_k^T, u_k^T)^T$, we aim to predict the state of the next step, $\hat{x}_{k+1}$. 
Because $\hat{x}_{k+1}$ is computed through GPR, it comes with a variance $\sigma^2$ that depends on the location of observed points, $z$. 
The lower the density of observations in the neighborhood of the inferred point, the larger the prediction variance. 
Hence, the variance can be seen as a proxy of the epistemic uncertainty of the model. 
We use such variance as our \mde\ value in Alg. \ref{alg: selection}--\ref{alg: expansion}.

\begin{algorithm}[tpb]
 \DontPrintSemicolon
  \SetKwFunction{SELECT}{SELECT}
  \SetKw{KwIn}{in}
  \SetKwProg{Fn}{Function}{:}{}  
  \Fn{\SELECT{$v$, $\tau$}}{
    \While{$v.\mathrm{isFullyExpanded}$ is true}{
        $\mathcal{V} \leftarrow \textt{getChildren}(v)$\;
        \RED{\For{$v_i \in \mathcal{V}$}{
            $\delta_i \leftarrow \frac{e^{\mde(v_i)/\tau}}{\sum_{v_j \in \mathcal{V}} e^{\mde(v_j)/\tau}}$\;
        }
        }
        $v \leftarrow \! \argmax_{v_i \in \mathcal{V}} \Big( \frac{Q(v_i)}{N(v_i)} + c \sqrt{\frac{\ln{N(v)}}{N(v_i)}} \Big) \RED{(1-\delta_i)}$\;
    }
    \KwRet $v$\;
  }
\caption{Uncertainty-aware selection function}
\label{alg: selection}
\end{algorithm}

\begin{algorithm}[tpb]
 \DontPrintSemicolon
  \SetKwFunction{EXPAND}{EXPAND}
  \SetKw{KwIn}{in}
  \SetKwProg{Fn}{Function}{:}{} 
  \Fn{\EXPAND{$v$, $h$}}{
    $\mathcal{A} \leftarrow \textt{getPossibleActions}(v)$\;
    $\mathcal{V} \leftarrow \emptyset$\;
    \For{$a_i \in \mathcal{A}$}{
            $x \leftarrow f(v.x, a_i)$\;
            create a node $v_i$ with state $v_i.x \leftarrow x$\;
            $\mathcal{V} \leftarrow \mathcal{V} \cup \{v_i\}$ \;
        }
    \RED{$\theta = \frac{1}{|\mathcal{V}|}\sum_{v_i \in \mathcal{V}} \mde(v_i) $\;
    \For{$v_i \in \mathcal{V}$}{
        $p' \leftarrow \textt{uniform}(0,1)$\;
        $p'' \leftarrow \frac{1}{1+e^{h(\mde(v_i) - \theta) }}$\;
        \If{$p' > p''$}{
              add node $v_i$ to the tree\;
        }   
    }
    }
    $v.\mathrm{isFullyExpanded} \leftarrow \textt{True}$\;
    \KwRet a random child of $v$\;
  }
\caption{Uncertainty-aware expansion function}
\label{alg: expansion}
\end{algorithm}

\section{Application to Robotized Pouring} \label{sec: case-study}

We consider a robotized liquid pouring case study and show how our uncertainty-aware approach can be applied to increase the success rate during execution. 
This section describes the pouring problem, the experimental setup, and the modeling approach.

\subsection{Problem definition} \label{sec: pouring-problem-definition}

The pouring problem consists of filling an output container with a sequence of pouring actions. 
We assume the liquid level can only be measured after executing each action, which is typical when using in-hand cameras.

The system state, $x$, is the liquid's level in the output container, measured as a percentage of the container's maximum fill level. 
Each action consists of a robot's tool rotation, $\alpha$, for a time duration $d$, so that $u_k = (\alpha_k, d_k)^T$. 
The goal is to reach a reference filling percentage, $x_{\mathrm{ref}}$, with a tolerance of $\pm c$, i.e., $X_{\mathrm{goal}} = [x_{\mathrm{ref}}-c, x_{\mathrm{ref}}+c]$.
We use GPR to predict the level, given the current level and a new action, so that $\hat{x}_{k+1} = \hat{f}(x_k,u_k)$. 

We use the following sparse reward function $r$ (i.e., only terminal states receive a reward):
\begin{equation}
    r(x_k,u_k) = \begin{cases}
        \rho(x_k,u_k), & \text{if $\textt{isTerminal}(x_k,u_k)=\textt{True}$}\\
        0, & \text{otherwise}
     \end{cases}
\end{equation}
\begin{equation}
    \rho(x_k,u_k) = \begin{cases}
			1+\frac{1}{k+1}, & \text{if $\hat{f}(x_k,u_k) \leq x_{\mathrm{ref}}+c$}\\
            0, & \text{otherwise}
		 \end{cases}
\end{equation}
\begin{equation}
    \textt{isTerminal}(x_k,u_k) = \begin{cases}
        \textt{True}, & \text{if $\hat{f}(x_k,u_k) \geq x_{\mathrm{ref}}-c$}\\
        &  \,\,\,\text{or } k>N_{\mathrm{max}} \\
        \textt{False}, & \text{otherwise}
     \end{cases}
\end{equation}
where $N_{\mathrm{max}}$ is the maximum search depth.
The reward is greater than zero only for those plans that do not exceed the chosen level threshold, $x_{\mathrm{ref}}+c$.
Moreover, the reward is inversely proportional to the number of actions needed to reach the terminal state.

Finally, we discretize the action space with a step of 0.25 rad and 0.1 seconds for $\alpha$ and $d$, respectively.

\begin{figure*}[tpb]
    \centering
    \subfloat[][Crop raw RGB]{
        \includegraphics[height=0.15\linewidth]{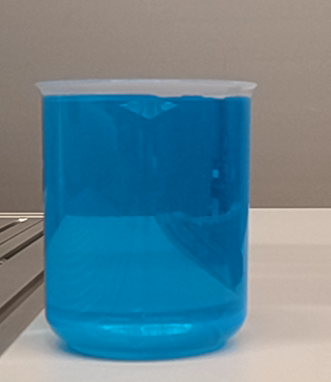}
        \label{fig:RGB_crop}
    }
    \,
    \subfloat[][Add Gaussian blur]{
        \includegraphics[height=0.15\linewidth]{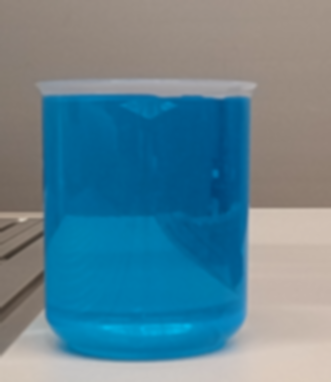}
        \label{fig:blur_crop}
    }
    \,
    \subfloat[][Convert to HSV]{
        \includegraphics[height=0.15\linewidth]{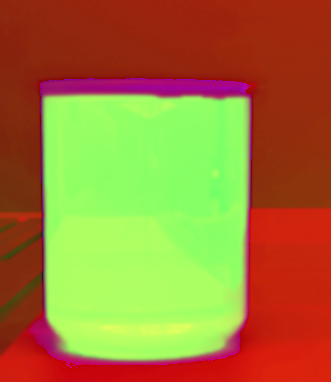}
        \label{fig:hsv_crop}
    }
    \,
    \subfloat[][Apply thresholding and find blob]{
        \includegraphics[height=0.15\linewidth]{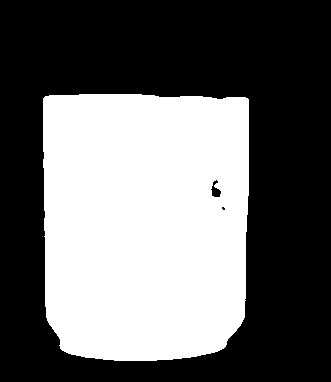}
        \label{fig:countour_crop}
    }
    \,
    \subfloat[][Compute level from aspect ratio]{
        \includegraphics[height=0.15\linewidth]{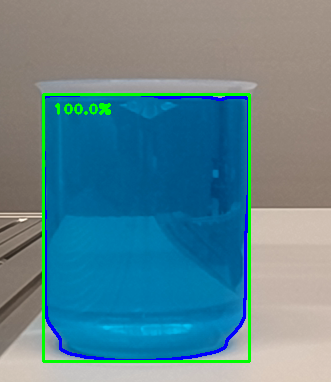}
        \label{fig:level_measured}
    }
    \captionimg{Image processing pipeline from RGB image to liquid volume identification.}
    \label{fig:image-processing-pipeline}
    \vspace{-0.3cm}
\end{figure*}


\subsection{Experimental setup} \label{sec: setup}

The setup is in Fig. \ref{fig: setup}. 
We use a 6-dof manipulator, Universal Robots UR5e, with a 3D-printed custom tool to hold a standard wine bottle. 
The robot is controlled using ROS Noetic through the \emph{MoveIt!} library \cite{moveit}. 

The perception system consists of a Luxonis OAK-D Pro RGB-D camera. 
The perception pipeline uses the RGB image from the camera and applies the following steps: 
(i) It crops the RGB image in a region of interest containing the container; 
(ii) It applies Gaussian blurring; 
(iii) It converts the image to HSV coordinates and thresholds it to isolate the hue of the liquid.
(iv) It extracts the larger resulting blob and computes its bounding box;
(v) Based on the aspect ratio of the bounding box, $\mathrm{ar}$, and knowing the aspect ratio of the bounding box of the full container, $\mathrm{ar}_{\mathrm{full}}$, it computes the filling percentage as
$l = 100\, \mathrm{ar}  / \mathrm{ar}_{\mathrm{full}}\,\, [\%]
$.
%

The pipeline is illustrated in Fig. \ref{fig:image-processing-pipeline}. 
Note that, to facilitate perception, we use dyed water and a transparent output container. 
With this perception pipeline, we were able to achieve a measurement accuracy of around 1\%. 

\subsection{Pouring modeling} \label{sec: gp-experiments}

We used GPR to model the pouring dynamics. 
We modeled the pouring process with different dataset sizes (40, 20, 10, and 5 points).
Each dataset point is a pouring and consists of the current level, $x_k$, and the action to be performed, $u_k = (\alpha_k, d_k)^T$.
The target value is the next level, $x_{k+1}$.
Points in the 40-point dataset were chosen randomly.
The smaller datasets were created by randomly sampling points from the 40-point dataset.
After a kernel selection phase, we opted for the combination of a Dot Product and a Rational Quadratic kernel.
Table \ref{tab: mse-gp} shows the Mean Squared Error (MSE) of the GPR models computed on a test set of 20 points.
As expected, the smaller the dataset, the greater the MSE value, meaning predictions are likely to be less accurate when the training dataset is smaller.


\section{Experiments} \label{sec: experiments}

\begin{table}[tpb]
    \captionimg{Mean Squared Error of GPR models. 
	}	
	\label{tab: mse-gp}
	\centering
	$
	\begin{array}{lcccc}
	\toprule
	\text{Dataset size} & \text{40 points} & \text{20 points} & \text{10 points} & \text{5 points}  \\
    \midrule
	\text{MSE}  &  18.6 & 24.5 & 25.3 & 33.4  \\
	\bottomrule
	\end{array}
	$
 \vspace{-0.2cm}
\end{table}

This section compares different planning algorithms, showing that uncertainty-aware methods can carry out pouring tasks with a higher success rate.

\subsection{Tests and metrics} \label{sec: metrics}


We compare the following algorithms:
\begin{itemize}
    \item MCTS: A standard MCTS algorithm without uncertainty awareness.
    \item UA-MCTS-0: The uncertainty-aware MCTS algorithm  proposed in \cite{ua_mcts}.
    \item MCTS-inflated: Standard MCTS with a conservative model propagation, i.e., we compute $\hat{x}_{k+1}=\hat{f}(x_k,u_k) + w\,\mde(x_k,u_k)$ during the search. This is a hand-crafted robust baseline that overestimates the effect of each pouring action proportionally to the uncertainty of transitions.
    \item UA-MCTS-1: The proposed approach with selection and expansion functions described in Alg. \ref{alg: selection}-\ref{alg: expansion}.
\end{itemize}
All methods use the same GPR models, reward functions, terminal-state checkers, and computational time (0.5 s).

We performed 30 tests for each model and method and evaluated the success rate (\% of tests where the final level belongs to $X_{\mathrm{goal}}$) and the number of actions to reach a terminal state. 
We define $X_{\mathrm{goal}} = [x_{\mathrm{ref}}-c, x_{\mathrm{ref}}+c$], where $c=2.5\%$ and  $x_{\mathrm{ref}}$ is chosen randomly (the same value of $x_{\mathrm{ref}}$ is used for all four methods).

The sigmoid steepness, $h$, and the softmax temperature, $\tau$, in Alg. \ref{alg: selection}-\ref{alg: expansion}, and the inflation factor $w$ in MCTS-inflated are tunable parameters. 
The inflation factor $w$ in MCTS-inflated determines how conservative the model propagation is. 
Large values of $w$ will greatly overestimate the effect of pouring actions, leading to many small pourings. 
Vice versa, if $w$ is close to one, MCTS-inflated will behave similarly to MCTS.
We used $h=10$, $\tau=0.1$, and $w=2$ for all tests.
We leave the quantitative analysis of the effects of parameter tuning on the results of the method as future work.

\subsection{Results} \label{sec: results}

\begin{table}[tpb]
    \captionimg{Experimental results.}	
    \label{tab: exp-results}
    \centering
    \makebox[0.9\linewidth]
    {
    	\centering
    	\subfloat[][40-point dataset]
    	{
    		$
            \begin{array}{lcc}
            \toprule
            & \text{Success rate} & \text{N. of actions }  \\
            & \text{[\%]} & \text{\footnotesize mean (std.dev.) }  \\
            \midrule
            \text{MCTS}  &  80 & 1.45(0.59)   \\
            \text{UA-MCTS-0} &  77 & 1.45(0.59)   \\
            \text{MCTS-inflated}  & \textbf{100} & 1.70(0.95) \\
            \text{UA-MCTS-1}  &  \textbf{100} & 2.05(1.02)  \\
            \bottomrule
            \end{array}
    		$
    		\label{tab: exp-res-40}
    	}
     }\\
     \makebox[0.9\linewidth]
     {
    	\subfloat[][20-point dataset]
    	{
    		$
    		\begin{array}{lcc}
            & \text{Success rate} & \text{N. of actions }  \\
            & \text{[\%]} & \text{\footnotesize mean (std.dev.) }  \\
            \midrule
        	\text{MCTS}  &  66 & 1.80(0.87)   \\
        	\text{UA-MCTS-0}  &  66 & 1.80(0.87)    \\
        	\text{MCTS-inflated}  & 93 & 2.55(1.20)  \\
        	\text{UA-MCTS-1}  &  \textbf{97} & 2.50(1.56)  \\
        	\bottomrule
        	\end{array}
    		$
    		\label{tab: exp-res-20}
    		}
    }\\
    \makebox[0.9\linewidth]
    {
    	\centering
    	\subfloat[][10-point dataset]
    	{
    		$
        	\begin{array}{lcc}
        	\toprule
            & \text{Success rate} & \text{N. of actions }  \\
            & \text{[\%]} & \text{\footnotesize mean (std.dev.) }  \\
            \midrule
        	\text{MCTS}  &  60 & 1.65(1.01)   \\
        	\text{UA-MCTS-0}  &  60 & 1.65(1.01)   \\
        	\text{MCTS-inflated}  &  97 & 2.60(1.07)  \\
        	\text{UA-MCTS-1}  &  \textbf{100} & 2.75(1.48)  \\
        	\bottomrule
        	\end{array}    
    		$
    		\label{tab: exp-res-10}
    	}
    }\\
    \makebox[0.9\linewidth]
    {
        \subfloat[][5-point dataset]
    	{
    		$
        	\begin{array}{lcc}
        	\toprule
            & \text{Success rate} & \text{N. of actions }  \\
            & \text{[\%]} & \text{\footnotesize mean (std.dev.) }  \\
            \midrule
        	\text{MCTS}  &  64 & 1.20(0.40)   \\
        	\text{UA-MCTS-0}  &  64 & 1.20(0.40)   \\
        	\text{MCTS-inflated}  &  \textbf{97} & 2.85(0.85)  \\
        	\text{UA-MCTS-1}  &  \textbf{97} & 2.40(1.16)  \\
        	\bottomrule
        	\end{array}
    		$
    		\label{tab: exp-res-5}
    	}
    }
    \vspace{-1cm}
\end{table}

Table \ref{tab: exp-results} shows the success rate and the number of actions to reach the goal for all methods and underlying models.
MCTS-inflated and UA-MCTS-1 have comparable success rates across all models and are significantly better than MCTS and UA-MCTS-0.
This difference becomes wider as the size of the training dataset decreases (up to 40\% for the 10-point model).
This fact is expected for MCTS because the model becomes less accurate and MCTS does not take the GP's variance into account during the search. 
Unexpectedly, we did not observe a significant difference between MCTS and UA-MCTS.
Possibly, the selection and expansion strategies proposed in \cite{ua_mcts} are not biased enough to discard inaccurate actions for the problem at hand.

Even though MCTS-inflated and UA-MCTS have similar success rates, the number of actions they take to reach the goal differs significantly.
With more accurate models (e.g., 40-point GP), MCTS-inflated has a number of actions comparable with standard MCTS and fewer than UA-MCTS-1 (see rows 1 and 3 of Table \ref{tab: exp-res-40}).
As the model becomes less accurate (e.g., 5-point GP), MCTS-inflated's number of actions is the highest among all methods (see rows 1 and 3 of Table \ref{tab: exp-res-5}).
This behavior is explained by considering the propagation function used in MCTS-inflated: by inflating the predicted level, the planner tends to predict higher levels when the variance of the action is large.
This does not prevent it from taking inaccurate actions, yet their effect is greatly overestimated.
As a consequence, the actual level will be lower than the predicted one and the number of actions required to reach the goal will increase, especially when the average variance estimated by the GP is large.
The UA-MCTS-1 approach is different: it chooses actions with low variance, no matter the average accuracy of the model. 
As a consequence, the algorithm may take more actions to reach the goal (see last rows of Tables \ref{tab: exp-res-40}-\ref{tab: exp-res-5}).

\begin{figure}[tpb]
\centering
    \centering
	\includegraphics[trim = 2.6cm 9cm 18cm 2.5cm, clip, angle=0, width=0.9\columnwidth]{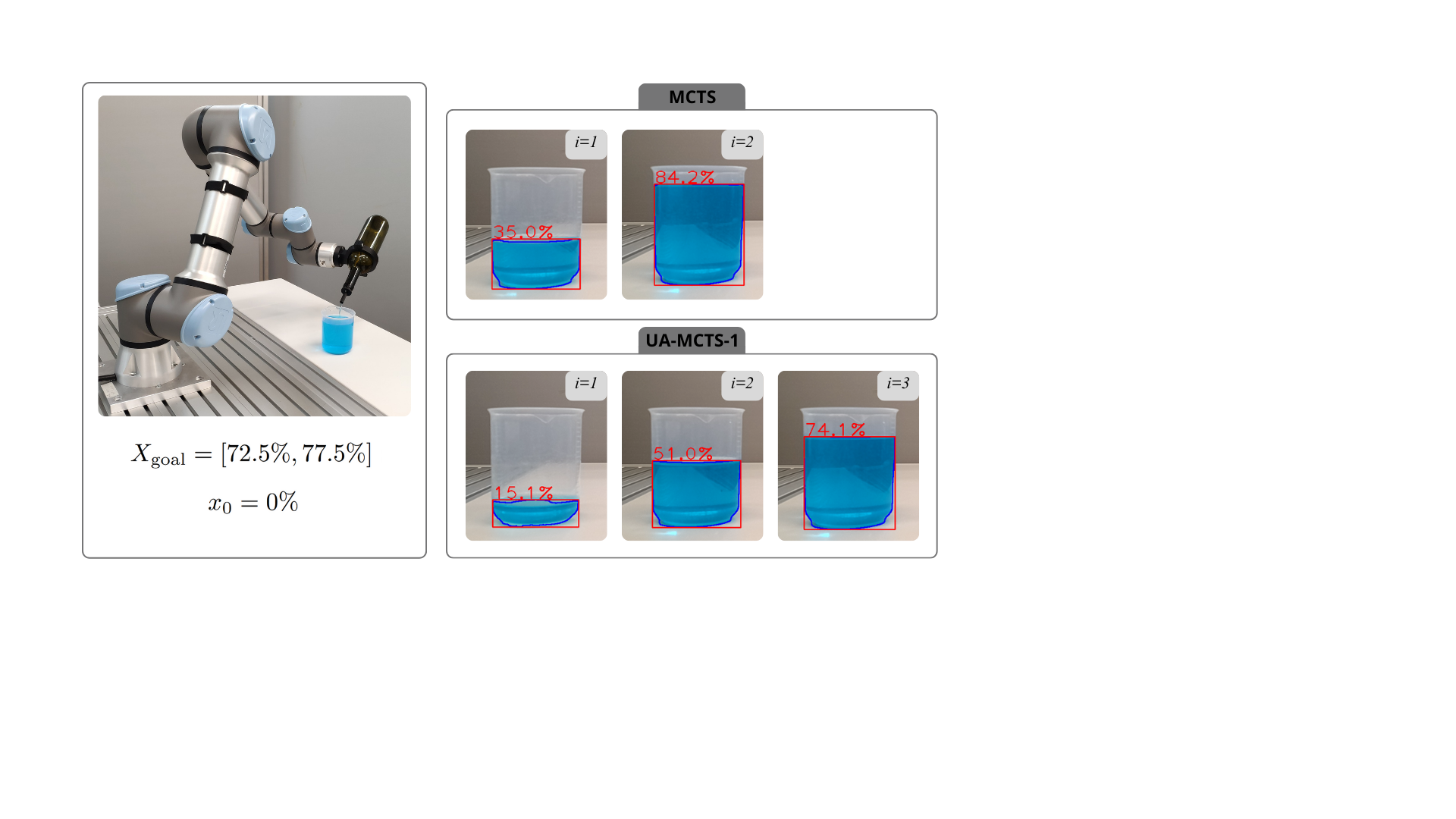}
	\captionimg{Example of pouring sequences with uncertainty-aware and -unaware methods.}
    \label{fig: exp-images}
    \vspace{-0.1cm}
\end{figure}

Examples of action sequences generated by MCTS and UA-MCTS-1 are in Fig. \ref{fig: exp-images}. MCTS tends to reach the goal with fewer actions, yet exceeding the goal level, leading to a failure. 
By contrast, UA-MCTS-1 is more conservative and takes one action more, yet it correctly reaches the goal. 
Please refer to the accompanying video for more details.

\begin{figure}[tpb]
\centering
    \centering
	\includegraphics[trim = 0cm 0cm 0cm 2cm, clip, angle=0, width=\columnwidth]{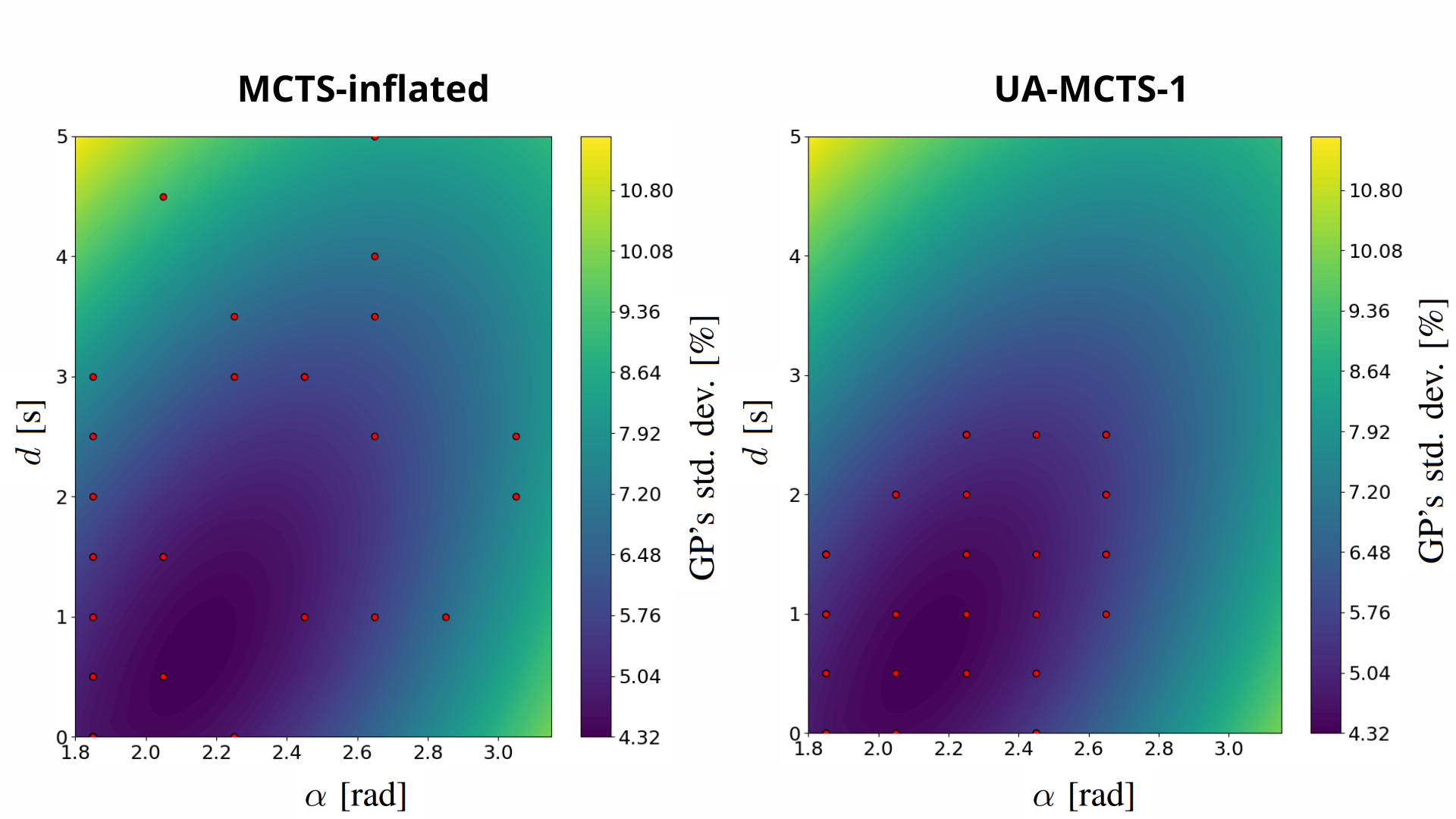}
	\captionimg{Actions $(\alpha,d)$ taken by the algorithm superimposed on the standard deviation of the 5-point GP model.}
    \label{fig: exp-gp-variance}
    \vspace{-0.4cm}
\end{figure}

Fig. \ref{fig: exp-gp-variance} clarifies the different policies of the methods.
The actions of MCTS-inflated and UA-MCTS-1 are superimposed on the GP's standard deviation across the action space.
MCTS-inflated's actions are scattered around the action space independent of the action's variance (similar distributions were obtained for MCTS and UA-MCTS-0 but are not shown here for the sake of brevity).
By contrast, UA-MCTS-1's actions are always in the low-variance region of the action space (blue area in Fig. \ref{fig: exp-gp-variance}), demonstrating that the approach tends to discard high-variance actions.

Although MCTS-inflated and UA-MCTS-1 have comparable success rates, note that MCTS-inflated is less general than UA-MCTS-1.
Indeed, MCTS-inflated assumes that inflating the effect of an action produces a conservative effect during execution. 
This may be true for one-dimensional problems like the pouring task at hand but does not hold in general (especially for higher state dimensionality).
Conversely, UA-MCTS-1 generalizes to higher dimensions, as its strategy directly reasons about transitions' inaccuracy. 

\rev{Finally, note that all methods were tested with the same allotted computational time (0.5 s). 
The proposed method only requires additional computation to calculate a number of sigmoid and softmax functions proportional to the branching factor of the problem. 
This additional computation is typically negligible compared to the model propagation.
}

\section{Conclusions}\label{sec:conclusions}

We proposed an uncertainty-aware Monte Carlo Tree Search approach to reason through the prediction error of learned models in robotics tasks.
We use uncertainty estimates such as the variance of Gaussian Processes to bias the search toward low-uncertainty states and actions.
%
We demonstrated the approach in a robotized pouring task with an RGB-based perception system and a robot arm. 
Our approach proved robust even with GPs obtained from very small datasets (up to 5 points), while baseline approaches suffered a decrease in the execution success rate as the model became less accurate.
Future works will extend the validation to more complex scenarios, with different container shapes and including robot motion planning and collision avoidance. 
\rev{A quantitative analysis of the hyperparameter tuning effects will also be carried out.}


\bibliographystyle{IEEEtran}
\bibliography{bib,bib_new}

\end{document}